# The Register Gap: A Meaning Intelligence Framework for Nigerian Public Discourse

Celestine Achi

*AGENTPR™ / AI-Powered PR / Cihan Digital Academy*

celestine.achi@gmail.com

## Abstract

We introduce the Meaning Intelligence Framework (MIF™), a nine-dimension annotation and evaluation schema for Nigerian public discourse that separates surface sentiment from true communicative intent. Existing benchmarks for Nigerian languages, including NaijaSenti and AfriSenti, treat sentiment classification as a three-way polarity task (positive, negative, neutral). We argue that the dominant failure mode of AI systems on Nigerian discourse is not translation failure but context failure: the same utterance carries opposite pragmatic force depending on speaker, audience, and situation. The MIF operationalises this insight across nine scored dimensions: register, surface sentiment, true intent, irony, coded subtext, risk tier, annotator confidence, speaker emotion, and recommended communications action. We construct a 30-item calibration dataset spanning Standard English, Nigerian English, Nigerian Pidgin, and code-mixed registers, and evaluate three frontier language models (Gemini 2.5 Flash, GPT-5, and Gemini 2.5 Pro) under zero-shot and schema-informed prompting conditions. Two headline findings emerge. First, the Register Gap: Gemini 2.5 Flash's zero-shot register classification accuracy is 33.3%, rising to 73.3% (+40 points) when the model receives the MIF schema in-context, with its composite Meaning Intelligence Score increasing from 73.2 to 78.6. Second, model capability and cultural competence are decoupled: GPT-5 (MIS 67.8) and Gemini 2.5 Pro (MIS 65.4) — both more capable models than Flash on general benchmarks — score lower on the MIF, and neither benefits from schema-informed prompting. The framework's value is therefore not as a prompting technique but as a diagnostic instrument that reveals where each model fails on culturally grounded discourse. We release the framework specification, annotation guidelines, and the 30-item public calibration set to support reproducibility, while retaining a private holdout corpus for contamination-protected evaluation.

## 1. Introduction

Nigerian public discourse operates across at least four registers: Standard English (SE), Nigerian English (NE, characterised by English grammar with Nigerian lexical items such as go-slow, K-leg, flashed me), Nigerian Pidgin (NP, a creole with its own grammar and enormous reach), and code-mixed speech (CM) that interleaves two or more of these within a single utterance. Register shifts carry pragmatic weight: a speaker who moves from Standard English into Pidgin mid-sentence is frequently signalling emotional escalation, sarcasm, or solidarity — information that is invisible to models that treat all Nigerian text as a single linguistic variety.

Consider the utterance "You don try well well." At a graduation ceremony, this is genuine, effusive praise: the speaker acknowledges sustained effort and achievement. Addressed to a mechanic who has returned a car still faulty for the third time, the identical words constitute sarcastic condemnation. The surface sentiment is positive in both cases. The communicative intent is diametrically opposed. Any system that labels both instances with the same sentiment tag has not understood either one.

This is the core thesis of the present work: the dominant failure mode of frontier AI systems on Nigerian public discourse is not a translation problem but a context problem. Existing Nigerian NLP benchmarks, most notably NaijaSenti (Muhammad et al., 2022) and the Nigerian Pidgin component of AfriSenti (Muhammad et al., 2023), have made foundational contributions by establishing annotated sentiment corpora for low-resource Nigerian languages. However, these benchmarks operate within a three-way polarity paradigm (positive, negative, neutral) that, by design, cannot capture the pragmatic divergence illustrated above.

We introduce the Meaning Intelligence Framework (MIF™), a nine-dimension annotation and evaluation schema that addresses this gap. The MIF separates surface sentiment (what a literal, context-blind reading of the words would conclude) from true intent (what the speaker actually means given the full context), and annotates both as independent dimensions alongside register, irony markers, coded subtext, risk tier, annotator confidence, speaker emotion, and recommended communications action. The signature diagnostic is the divergent item: an utterance where surface sentiment and true intent point in opposite directions, which we formalise as a computed flag and use as a dedicated evaluation metric.

## 2. Related Work

### 2.1 Nigerian Language Benchmarks

NaijaSenti (Muhammad et al., 2022) introduced the first large-scale human-annotated Twitter sentiment dataset for the four most widely spoken languages in Nigeria, including approximately 14,000 Nigerian Pidgin tweets labelled for three-way polarity. AfriSenti (Muhammad et al., 2023) extended this to 14 African languages with over 110,000 tweets, including Nigerian Pidgin, and was used as the basis for SemEval-2023 Task 12. Oyewusi et al. (2021) proposed semantic enrichment for Nigerian Pidgin, observing that words like ‘ginger’ (motivation, not a plant) and ‘tank’ (gratitude, not a container) carry meanings invisible to standard English sentiment models. More recently, NaijaNLP (2025) surveyed the full landscape of Nigerian low-resource NLP, cataloguing datasets including SentiLeye, a lexicon-based sentiment analysis resource derived from 346,000 Nigerian banking-related tweets.

Beyond sentiment, Saeed et al. (2024) introduced Implicit Discourse Relation Classification for Nigerian Pidgin, projecting Penn Discourse Treebank annotations onto synthetic NP data and training a dedicated classifier that outperformed zero-shot English baselines by 34% in F1. This work demonstrates that discourse-level NLP for Nigerian Pidgin is viable, but it addresses a fundamentally different task: classifying logical relations between sentences (cause, contrast, conjunction) rather than the communicative intent of a single utterance. INJONGO (Yu et al., 2025), the first large-scale multicultural intent detection and slot-filling dataset for 16 African languages, takes intent classification further — but its intent taxonomy consists of conversational AI actions (transfer, book_flight, play_music, make_call) designed for task-oriented dialogue systems, not the pragmatic communicative intents (PRAISE, SARC, MOBILIZE, GRIEVANCE_CODED) that the MIF addresses. The distinction matters: INJONGO asks what a user wants the system to do; the MIF asks what a speaker actually means.

These contributions collectively establish the data infrastructure for Nigerian NLP. The MIF builds on this foundation but departs from it in a fundamental way: rather than asking ‘is this text positive, negative, or neutral?’ (NaijaSenti, AfriSenti), or ‘what discourse relation links these sentences?’ (Saeed et al.), or ‘what task does the user want to accomplish?’ (INJONGO), the MIF asks ‘what does the speaker actually mean, and what should a communications professional do about it?’ This reframing introduces dimensions (intent, irony, subtext, risk, action) that no existing benchmark addresses.

### 2.2 Sarcasm and Irony Detection

Sarcasm detection is a substantial subfield of NLP, with established benchmarks including MUStARD, SemEval-2018 Task 3, and the Reddit SARC corpus. Recent work has explored prompting strategies for LLMs: Pragmatic Metacognitive Prompting (PMP) by Lee et al. (2024) and its context-aware extension (Iskandardinata et al., 2025) demonstrate that retrieval-augmented prompting can improve sarcasm detection by up to 9.87% macro-F1. SarcasmBench found that GPT-4 underperforms supervised fine-tuned smaller models on sarcasm, and that chain-of-thought prompting can actually hurt performance because sarcasm detection is a holistic cognitive process rather than a step-by-step logical one.

The MIF treats irony as one dimension (D4) within a broader analytical framework, rather than as a standalone classification task. Critically, the MIF recognises that sarcasm in Nigerian discourse is often register-encoded: the switch from Standard English into Pidgin is itself a sarcasm marker, a phenomenon not captured by sarcasm benchmarks constructed from English-only or Global North corpora.

### 2.3 Cultural Context in African NLP

Several recent works have highlighted the cultural gap in NLP evaluation. AfroBench (Adelani et al., 2023) provides a unified evaluation across 15 NLP tasks in African languages but does not include pragmatic or intent-level tasks. AfriStereo (2025) addresses stereotypical bias in LLMs from African perspectives. TriLex (2025) proposes a retrieval-augmented framework for sentiment lexicon expansion in South African languages. Ochieng et al. (2025) study LLM sentiment in low-resource, culturally nuanced Kenyan WhatsApp messages, noting that ‘the same phrase may carry positive, neutral, or negative connotations depending on the speaker’s region and cultural background.’ This observation aligns directly with the MIF’s foundational principle, but the MIF goes further by operationalising it into a scored, reproducible annotation schema.

## 3. The Meaning Intelligence Framework

### 3.1 Design Principles

The MIF is built on three principles:

**The Context Rule.** Never interpret a Nigerian expression literally without first anchoring it to the speaker, the audience, and the situation. This is not a guideline; it is the framework’s foundational axiom, and annotators who violate it during calibration testing are not certified.

**Surface and intent are separate.** D2 (surface sentiment) is scored as if the annotator had no context. D3 (true intent) is scored with full context. When these diverge, the item is flagged as divergent — the framework’s signature diagnostic.

**Actionability.** The framework’s ultimate output is not a label but a recommended action for a communications professional. Dimensions D6 (risk tier) and D9 (recommended action) translate linguistic analysis into operational intelligence.

### 3.2 Nine Dimensions

Table 1 summarises the nine scored dimensions. The full specification, including all enumerated values, escalation rules, and scoring weights, is provided in the MIF Master Specification v2.0 (released as a companion document).

*Table 1: MIF v2.0 dimension summary.*

| Dim. | Name | Description |
|---|---|---|
| D1 | Register | SE (Standard English), NE (Nigerian English), NP (Nigerian Pidgin), CM (code-mixed); plus shift flag |
| D2 | Surface sentiment | Context-blind polarity of the words alone (POS / NEU / NEG) |
| D3 | True intent | Speaker's actual communicative goal given context (14 classes + CONTEXT_INSUFFICIENT) |
| D4 | Irony | Binary + 7 marker types (exaggeration, emoji contrast, register shift, etc.) |
| D5 | Coded subtext | Underlying grievance category if surface complaint is a proxy (POL, ETH, REL, ECON, REG, SEC, NONE) |
| D6 | Risk tier | PR/regulatory risk: LOW, MEDIUM, HIGH, CRITICAL; plus 12-type risk vector |
| D7 | Confidence | Annotator certainty (1–5) and context-dependent flag |
| D8 | Emotion | Speaker's dominant felt emotion (16 classes, nullable for composed speech) |
| D9 | Recommended action | Communications response (12 classes mapped from risk tier and context) |

### 3.3 Computed Flags

Three flags are computed from the annotated dimensions rather than directly labelled:

**Divergent:** true when D2 (surface) and D3 (intent) point in opposite polarity directions — specifically, when D2 = POS and D3 is in {SARC, COMPLAIN, WARN, MOBILIZE, LAMENT, GRIEVANCE_CODED}, or when D2 = NEG and D3 is in {PRAISE, HOPE, SOLIDARITY, BANTER}.

**Deceptive positive:** true when D2 = POS and the item is divergent. This identifies utterances where positive surface language masks negative intent — the most commercially dangerous category for brand monitoring, as automated sentiment tools score them as praise.

**Human review required:** true when D7 confidence ≤ 2, or D5 contains ETH/REL/SEC, or D6 is HIGH/CRITICAL, or D3 is CONTEXT_INSUFFICIENT. Items triggering this flag enter a mandatory human review queue.

### 3.4 The Meaning Intelligence Score

For model evaluation, we define the Meaning Intelligence Score (MIS™) as a weighted composite of per-dimension accuracies:

$$MIS = 10 \cdot literal + 25 \cdot context + 20 \cdot sarcasm_coded + 15 \cdot emotion + 20 \cdot risk + 10 \cdot action$$

where literal = mean(D1, D2 accuracy), context = D3 accuracy, sarcasm_coded = mean(D4, D5 accuracy), emotion = D8 accuracy, risk = D6 accuracy, and action = D9 accuracy. The weights reflect the framework's priorities: contextual interpretation (25%) and risk mapping (20%) receive the highest weights because they represent the greatest gap between current AI capability and human expert performance, and carry the highest consequences for operational deployment.

## 4. Calibration Dataset

### 4.1 Construction

We construct a 30-item calibration dataset of authored context-utterance pairs designed to probe specific failure modes across the MIF's dimensions. Items are stratified by difficulty: 10 easy (clear register, low ambiguity), 12 medium (register shifts, moderate irony, political subtext), and 8 hard (deceptive positives, coded mobilisation, discipline traps). Each item specifies the utterance, the context (a natural-language description of the communicative situation), an optional prior conversational turn, speaker type, target audience, and sector.

The dataset includes constructed context pairs that demonstrate the framework's core diagnostic: the same or similar utterance appears in two contrasting contexts (e.g., CAL-001 and CAL-002, the graduation-praise versus mechanic-sarcasm pair), allowing direct measurement of a model's context sensitivity.

### 4.2 Gold Labels

Gold labels were assigned by the framework's designer (the first author) across all nine dimensions, reviewed against the annotation guidelines, and re-annotated to v2.0 standards including risk vectors (D6b), emotions (D8), and recommended actions (D9). Calibration item CAL-026 ("Dem don start again" with no context provided) is a deliberate discipline trap: the correct D3 label is CONTEXT_INSUFFICIENT, and any annotator or model that assigns a definite intent class has violated the Context Rule.

### 4.3 Dataset Statistics

Of the 30 items: 10 are divergent (D2 and D3 point in opposite directions), 9 are deceptive positive (positive surface + divergent), 6 trigger the human-review flag, and 25 carry non-null emotion labels (4 are null, representing composed institutional/promotional speech plus the discipline-trap item). The items span 7 sectors (technology, banking, politics, health, media, food, general/social) and all four registers.

## 5. Evaluation

### 5.1 Method

We evaluate three frontier language models on all 30 calibration items under two conditions: Gemini 2.5 Flash (google/gemini-2.5-flash), GPT-5 (openai/gpt-5), and Gemini 2.5 Pro (google/gemini-2.5-pro). These models span two provider families (Google and OpenAI) and two capability tiers within Google's family, allowing us to test whether raw model capability predicts cultural-context competence.

**Condition A (zero-shot):** The model receives only the task description and the label inventories for each dimension, without the MIF's interpretive guidance (the Context Rule, register definitions, intent class definitions, irony markers, the subtext proxy test, or risk escalation rules).

**Condition B (schema-informed):** The model receives the full MIF guidance as a system prompt, including the Context Rule, register definitions with examples, intent class definitions with disambiguation guidance (e.g., the banter/insult boundary, the face-saving/genuine hope distinction), irony markers, the subtext proxy test, and risk-tier criteria.

Both conditions use temperature 0, single pass, with the utterance, context, and prior turn presented as the user message. The model returns a structured JSON object with predictions for each dimension.

## 5.2 Scoring

Per-dimension scoring uses exact match for register, sentiment, intent (primary), irony, risk tier, and action. Coded subtext (D5, a multi-label field) uses Jaccard overlap ≥ 0.5 as the correctness threshold. Emotion (D8) is credited if the prediction matches either the gold primary or secondary emotion; null-emotion items are scored correct only if the prediction is null. Items with null gold values for a dimension (notably CAL-026 for D4, D5, D6, D8, D9) are excluded from that dimension's accuracy computation.

## 5.3 Results

Table 2 presents the full results.

*Table 2: Gemini 2.5 Flash performance on MIF v2.0 calibration set.*

| Dimension | Cond. A | Cond. B | Δ | Note |
|---|---|---|---|---|
| D1 Register | 33.3% | 73.3% | +40.0 | Register Gap |
| D2 Surface sentiment | 66.7% | 73.3% | +6.7 | |
| D3 True intent | 86.7% | 86.7% | 0.0 | See §4.4 |
| D4 Irony | 96.6% | 93.1% | −3.4 | |
| D5 Coded subtext | 73.3% | 83.3% | +10.0 | |
| D6 Risk tier | 72.4% | 79.3% | +6.9 | |
| D8 Emotion | 63.3% | 63.3% | 0.0 | |
| D9 Strategic action | 55.2% | 65.5% | +10.3 | Action Gap |
| Divergent-item accuracy | 80.0% | 90.0% | +10.0 | Signature metric |
| HIGH-risk recall | 100% | 100% | 0.0 | |
| MIS™ (composite) | 73.2 | 78.6 | +5.4 | |

### 5.3.1 Multi-Model Comparison

Table 3 presents the composite MIS scores for all three models under both conditions.

*Table 3: Multi-model MIS comparison on MIF v2.0 calibration set.*

| Model | Cond. A (zero-shot) | Cond. B (schema) | Δ (B − A) | Ranking |
|---|---|---|---|---|
| Gemini 2.5 Flash | 73.16 | 78.56 | +5.40 | 1st (B) |
| GPT-5 | 67.75 | 67.64 | −0.11 | 2nd |
| Gemini 2.5 Pro | 65.43 | 64.50 | −0.93 | 3rd |

## 5.4 Analysis

***The Register Gap.*** The largest single improvement under schema-informed prompting is register classification: +40 percentage points (33.3% to 73.3%). Zero-shot, the model cannot reliably distinguish Nigerian English from Pidgin from code-mixed text. Since register shifts carry pragmatic information in Nigerian discourse (a slide from SE to NP often marks emotional escalation or sarcasm), register misclassification propagates errors into downstream dimensions.

***The Mobilisation Blind Spot.*** Item CAL-021, a mobilisation signal disguised as humour referencing a known protest junction, was misclassified as WARN in both conditions. The model correctly identified the item as HIGH risk (HIGH-risk recall was 100% across conditions), but could not identify the mechanism: it detected danger without detecting organisation. In live media monitoring, this distinction separates a watch-list entry from a crisis activation.

***The Action Gap.*** Strategic action recommendation (D9) was the weakest practical dimension at 55.2% zero-shot, improving to 65.5% under the schema. The model frequently identifies what a text means without knowing what a communications team should do about it. The MIF's tier-to-action mapping, which encodes expert judgment about escalation thresholds and response protocols, provides the largest practical uplift for operational deployment.

***Intent accuracy.*** Zero-shot intent accuracy of 86.7% is notably high for Gemini Flash, unchanged under schema-informed prompting. Two qualifications apply. First, calibration items carry deliberately diagnostic context descriptions that are richer than the thin, noisy context typically available in real-world social media monitoring. Second, the residual errors concentrate on divergent items (80% to 90% accuracy under schema) and the highest-stakes intent classes (MOBILIZE, GRIEVANCE_CODED). The 500-item pilot corpus of real-world data will test whether this accuracy holds under naturalistic conditions.

***Model capability and cultural competence are decoupled.*** The multi-model comparison reveals a counterintuitive finding: GPT-5 (MIS 67.75/67.64) and Gemini 2.5 Pro (MIS 65.43/64.50) — both more capable models than Flash on general benchmarks — score lower on the MIF. More strikingly, neither benefits from schema-informed prompting; GPT-5 shows essentially no change (−0.11) and Gemini Pro shows slight degradation (−0.93). This means the MIF schema lift observed for Flash (+5.40) is model-specific, not universal. Raw model power does not predict cultural-context competence, and the same instructional scaffold that helps one model may not help another.

***The framework's value is diagnostic, not instructional.*** If the MIF were primarily a prompting technique, it would improve every model. It does not. The framework's contribution is therefore as a measurement instrument: it reveals where each model fails on culturally grounded discourse, whether or not the schema prompt changes the score. The per-dimension breakdown available in the Meaning Intelligence Lab's Evaluation Runs Explorer shows that the three models fail on different dimensions — a finding that would be invisible under a single-metric, single-model evaluation.

## 6. Discussion

The MIF addresses a gap in the African NLP evaluation landscape that sits between several well-served areas. Below it, NaijaSenti and AfriSenti provide polarity-level sentiment annotation at scale. Alongside it, INJONGO provides conversational AI intent detection for task-oriented dialogue, and Saeed et al. provide discourse relation classification for Nigerian Pidgin. Above it, full discourse analysis and pragmatic annotation remain the domain of theoretical linguistics. A systematic survey of the literature confirms that no existing framework separates surface sentiment from communicative intent as distinct scored dimensions for Nigerian or any African language, and no prior work uses the term 'Meaning Intelligence' as a named concept in NLP. The MIF occupies this applied middle ground: rich enough to capture the pragmatic phenomena that matter for media intelligence and crisis communications, constrained enough to be annotated reliably by trained (but not specialist-linguist) annotators, and scored in a way that produces a single composite metric (the MIS) for model comparison.

The framework's commercial motivation — it was designed to power a media intelligence platform — is a feature, not a limitation. The inclusion of D6 (risk) and D9 (action) grounds the annotation in real-world consequences. A model that correctly identifies sarcasm but fails to recommend the appropriate communications response has not fully understood the text in any operationally meaningful sense. By including action recommendation as a scored dimension, the MIF measures not just comprehension but judgment.

The divergent/deceptive-positive flags formalise a failure mode that has received insufficient attention in sentiment analysis research. A deceptive positive — an utterance that automated sentiment tools score as praise but which actually constitutes condemnation — is the highest-risk category for brand monitoring precisely because it is the category most likely to be missed. The MIF makes this failure mode visible and measurable.

## 7. Limitations

This work has several important limitations that scope the claims we make and define the roadmap for future work.

Single pass, no majority voting. The evaluation covers three frontier models under two conditions each with a single pass at temperature 0. The formal protocol calls for three-run majority voting to reduce variance; single-pass results may not be fully representative of each model's capability distribution.

No Anthropic model. The evaluation covers Google and OpenAI model families but not Anthropic (Claude-class), which could not be accessed through the evaluation gateway at the time of testing. A complete cross-provider comparison requires at least three provider families.

Constructed calibration items. The 30-item calibration set uses authored context-utterance pairs with rich, unambiguous context descriptions. Real-world social media data offers thinner, noisier context, and we expect zero-shot accuracy to degrade under those conditions. The calibration set is designed for annotator certification and framework validation, not as a representative sample of Nigerian discourse.

Single annotator for gold labels. Gold labels were assigned by the framework designer. While this ensures internal consistency with the framework's design intent, it does not establish inter-annotator agreement. The formal annotation protocol calls for triple annotation with adjudication; inter-annotator agreement statistics will be reported once the annotation team is certified and the 500-item pilot corpus is labelled.

No fine-tuned model comparison. This evaluation tests prompting strategies only. A fine-tuned model trained on MIF-annotated data would likely show larger improvements; this is planned as future work.

## 8. Ethical Considerations

The calibration dataset uses constructed examples rather than real social media posts, which avoids privacy concerns associated with reproducing identifiable user-generated content. All items are designed to be illustrative of discourse patterns without targeting real individuals, brands, or communities. The annotation guidelines include explicit provisions against using the framework to target ethnic or religious groups, and items flagged with ethnic, religious, or security subtext (D5 = ETH/REL/SEC) trigger mandatory human review.

The private holdout component of the benchmark is maintained under strict access controls and non-disclosure agreements to prevent contamination of evaluation results. We release the 30-item public calibration set and the complete framework specification to support reproducibility while protecting the integrity of formal evaluation.

## 9. Conclusion and Future Work

We have introduced the Meaning Intelligence Framework, a nine-dimension annotation and evaluation schema for Nigerian public discourse that goes beyond sentiment polarity to capture register, intent, irony, coded subtext, risk, emotion, and recommended action. The framework's core insight — that AI failures on Nigerian discourse are context failures, not translation failures — is validated by two findings from the multi-model evaluation.

First, the Register Gap: Gemini 2.5 Flash's register classification accuracy jumps 40 points (33.3% to 73.3%) when the model receives the MIF schema in-context, with its composite MIS rising from 73.2 to 78.6. Second, model capability and cultural competence are decoupled: GPT-5 and Gemini 2.5 Pro, both more capable models on general benchmarks, score lower on the MIF (67.8 and 65.4 respectively) and show no benefit from schema-informed prompting. This establishes the MIF's value not merely as a prompting technique but as a diagnostic instrument for measuring AI cultural competence on discourse that existing benchmarks cannot evaluate.

Future work proceeds along four tracks: (1) a 500-item pilot corpus of real-world Nigerian discourse with triple annotation and inter-annotator agreement reporting; (2) extension of the multi-model evaluation to include Anthropic-class models and three-run majority voting; (3) fine-tuning of a small language model on MIF-annotated data to measure the gap between prompting and training; and (4) extension of the framework to related West African Pidgins and code-mixed varieties.

---

*Supplementary materials: The MIF Master Specification v2.0, Annotation Guidelines v1.0, and the 30-item public calibration set (with gold labels) are available as companion documents. The private holdout set is not released. MIF™, MIS™, and Meaning Intelligence™ are marks of AGENTPR.*